\documentclass[conference]{IEEEtran}
\IEEEoverridecommandlockouts
\usepackage{cite}
\usepackage{amsmath,amssymb,amsfonts}
\usepackage{algorithmic}
\usepackage{graphicx}
\usepackage{subcaption}
\usepackage{textcomp}
\usepackage{xcolor}
\usepackage{array}

\def\BibTeX{{\rm B\kern-.05em{\sc i\kern-.025em b}\kern-.08em
    T\kern-.1667em\lower.7ex\hbox{E}\kern-.125emX}}
\begin{document}

\title{Adiabatic Quantum Linear Regression \\
\thanks{This manuscript has been authored in part by UT-Battelle, LLC under Contract No. DE-AC05-00OR22725 with the U.S. Department of Energy. The United States Government retains and the publisher, by accepting the article for publication, acknowledges that the United States Government retains a non-exclusive, paid-up, irrevocable, world-wide license to publish or reproduce the published form of this manuscript, or allow others to do so, for United States Government purposes. The Department of Energy will provide public access to these results of federally sponsored research in accordance with the DOE Public Access Plan (http://energy.gov/downloads/doe-public-access-plan). This research used resources of the Oak Ridge Leadership Computing Facility, which is a DOE Office of Science User Facility supported under Contract DE-AC05-00OR22725.}
}

\author{\IEEEauthorblockN{Prasanna Date}
\IEEEauthorblockA{\textit{Computational Data Analytics} \\
\textit{Oak Ridge National Laboratory}\\
Oak Ridge, TN \\
datepa@ornl.gov}

\and
\IEEEauthorblockN{Thomas Potok}
\IEEEauthorblockA{\textit{Computational Data Analytics} \\
\textit{Oak Ridge National Laboratory}\\
Oak Ridge, TN \\
potokte@ornl.gov}

}

\maketitle

\begin{abstract}
A major challenge in machine learning is the computational expense of training these models. 
Model training can be viewed as a form of optimization used to fit a machine learning model to a set of data, which can take up significant amount of time on classical computers. 
Adiabatic quantum computers have been shown to excel at solving optimization problems, and therefore, we believe, present a promising alternative to improve machine learning training times. 
In this paper, we present an adiabatic quantum computing approach for training a linear regression model. 
In order to do this, we formulate the regression problem as a quadratic unconstrained binary optimization (QUBO) problem.
We analyze our quantum approach theoretically, test it on the D-Wave 2000Q adiabatic quantum computer and compare its performance to a classical approach that uses the Scikit-learn library in Python.
Our analysis shows that 
the quantum approach attains up to $\boldsymbol{2.8 \times}$ speedup over the classical approach on larger datasets, 
and performs at par with the classical approach on the regression error metric.
\end{abstract}


\section{Introduction}
\label{sec:intro}

Machine learning algorithms and applications are ubiquitous in our day-to-day lives and are deployed on a variety of devices---from edge devices like smartphones to large supercomputers.
Before they are deployed in a real world application, machine learning models need to be trained, which is a time intensive process, and can even take a few months.
When training machine learning models, we usually minimize a well defined error function and leverage optimization techniques like gradient descent, ellipsoid method, evolutionary optimization etc. \cite{date2019combinatorial}.
While these techniques work well on smaller problems, they become computationally demanding, time consuming and infeasible as the problem size increases.

Quantum computers are known to be good at solving hard optimization problems and offer a promising alternative to accelerate the training of machine learning models \cite{wittek2014quantum}.
For instance, adiabatic quantum computers like the D-Wave 2000Q can approximately solve NP-complete problems like the quadratic unconstrained binary optimization (QUBO) problem efficiently, and have been used to train machine learning models like Restricted Boltzmann Machines (RBMs) and Deep Belief Networks (DBNs) in classical-quantum hybrid approaches \cite{date2019classical}.
Although today's quantum computers are small, error-prone and in the noisy intermediate-scale quantum (NISQ) era, the future machines are sought to be large, reliable and scalable \cite{preskill2018quantum, ladd2010quantum}.

In this paper, we evaluate the use of adiabatic quantum computers to train linear regression models.
Linear regression is a machine learning technique that models the relationship between a scalar dependent variable and one or more independent variables \cite{montgomery2012introduction}.
It has applications in business, economics, astronomy, scientific analysis, weather forecasting, risk analysis etc. \cite{wu2002new, yatchew1998nonparametric, isobe1990linear, leatherbarrow1990using, glahn1972use, guikema2008flexible}.
It is not only used for prediction and forecasting, but also to determine the relative importance of data features.
Linear regression has an analytical solution and can be solved in $\mathcal{O}(N^3)$ time on classical computers, where $N$ is the size of the training data.
While seemingly efficient on smaller problems, the existing algorithms tend to become infeasible as the problem size grows despite being executed in parallel.
So, it is necessary to explore the applicability of non-conventional computing paradigms like quantum computing for linear regression.

The main contributions of this work are as follows:
\begin{enumerate}
    \item We propose a quantum approach to solve the linear regression problem by formulating it as a quadratic unconstrained binary optimization (QUBO) problem.
    \item We theoretically analyze our quantum approach and demonstrate that its run time is equivalent to that of current classical approaches.
    \item We empirically test our quantum approach using the D-Wave 2000Q adiabatic quantum computer and compare its performance to a classical approach that uses the Scikit-learn library in Python. The performance metrics used for this comparison are regression error and computation time. We show that both approaches achieve comparable regression error, and that the quantum approach achieves $2.8\times$ speedup over the classical approach on larger datasets.
\end{enumerate}


\section{Related Work}
\label{sec:related}

Linear regression is one of the most widely used statistical machine learning techniques.
Bloomfield and Steiger propose a method for least absolute deviation curve fitting, which was three times faster than the ordinary least squares approach \cite{bloomfield1980least}. 
Megiddo and Tamir propose $\mathcal{O}(N^2 \log N)$ and $\mathcal{O}(N \log^2 N)$ algorithms for regression based on the Euclidean error and the rectilinear ($l_1$) error respectively, where $N$ is the number of datapoints in the training dataset \cite{megiddo1983finding}.
Zemel propose $\mathcal{O}(N)$ algorithm for linear multiple choice knapsack problem, which translates to linear regression with rectilinear error \cite{zemel1984n}.

Theoretically, the best classical algorithm for linear regression, has time complexity $\mathcal{O}(N^2 \log N)$, where $N$ is the number of datapoints in the training dataset.
However, most practical implementations in widely used machine learning libraries like the Scikit-learn library in Python run in $\mathcal{O}(N d^2)$ time, where $d$ is the number of features in the training dataset \cite{pedregosa2011scikit, buitinck2013api}.
$\mathcal{O}(N d^2)$ appears to be the most widely accepted time complexity for linear regression, and will be the basis of comparison in this paper.

Quantum algorithms have also been explored for linear regression in the literature.
Harrow et al. propose a quantum algorithm for solving a system of linear equations, that runs in $\texttt{poly}(\log N, \kappa)$ time, where $\kappa$ is the condition number of the input matrix \cite{harrow2009quantum}.
Schuld et al. propose an algorithm for linear regression with least squares that runs in logarithmic time in the dimension of input space provided training data is encoded as quantum information \cite{schuld2016prediction}.
Wang proposes a quantum linear regression algorithm that runs in $\texttt{poly}(\log_2 N, d, \kappa, \frac{1}{\epsilon})$, where $\epsilon$ is the desired precision in the output \cite{wang2017quantum}.
Dutta et al. propose a 7-qubit quantum circuit design for solving a 3-variable linear regression problem and simulate it on the Qiskit simulator \cite{dutta2018demonstration}.
Zhang et al. propose a hybrid approach for linear regression that utilizes both discrete and continuous quantum variables \cite{zhang2019realizing}.

Adiabatic quantum computers have also been used to address machine learning problems in limited capacity.
Foster et al. explore the use of D-Wave quantum computers for statistics \cite{foster2019applications}.
Djidjev et al. use the D-Wave 2X quantum annealer for combinatorial optimization \cite{djidjev2018efficient}.
Borle et al. present a quantum annealing approach for the linear least squares problem \cite{borle2019analyzing}.
Chang et al. propose a quantum annealing approach for solving polynomial systems of equations using least squares \cite{chang2019least}.
Chang et al. present a method for solving polynomial equations using quantum annealing and discuss its application to linear regression \cite{chang2019quantum}.
Neven et al. train a binary classifier with the quantum adiabatic algorithm and show that it performs better than the state-of-the-art machine learning algorithm AdaBoost \cite{neven2008training}.
Adachi and Henderson use quantum annealing for training deep neural networks on the coarse-grained version of the MNIST dataset \cite{adachi2015application}.
Date et al. propose a classical quantum hybrid appraoch for unsupervised probabilistic machine learning using Restricted Boltzmann Machines and Deep Belief Networks \cite{date2019classical}.

While several quantum computing approaches have been proposed for linear regression, most of them leverage universal quantum computers and not adiabatic quantum computers.
Moreover, they have not been empirically validated on real hardware to the best of our knowledge.
In this work, we propose a quantum computing approach for linear regression that leverages adiabatic quantum computers, which have shown to be much more scalable than universal quantum computers in the recent past.
Furthermore, we empirically validate our approach on synthetically generated datasets.

\section{Linear Regression}
\label{sec:background}

\begin{figure}
    \centering
    \includegraphics[scale=0.5]{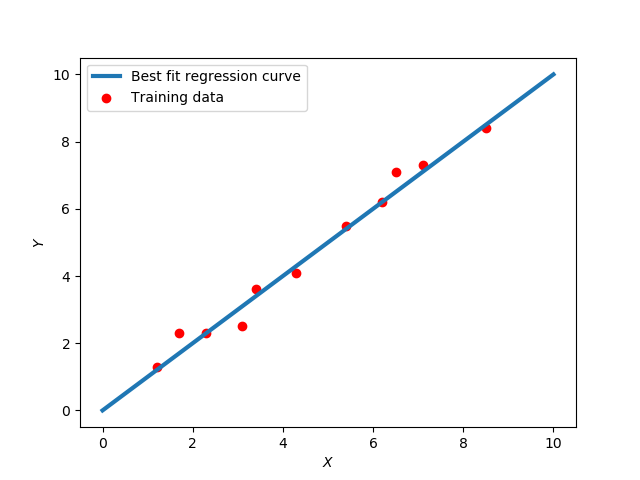}
    \caption{Linear Regression. Red dots represent the training data for regression, and blue line represents the best fit for the given training data.}
    \label{fig:regression-explained}
\end{figure}

We use the following notation throughout this paper:
\begin{itemize}
    \item $\mathbb{R}$: Set of real numbers
    \item $\mathbb{B}$: Set of binary numbers, i.e. $\mathbb{B} = \{0, 1\}$.
    \item $\mathbb{N}$: Set of natural numbers
    \item $X$: Augmented training dataset, usually $X \in \mathbb{R}^{N \times (d+1)}$, i.e. $X$ contains $N$ data points ($N \in \mathbb{N}$) along its rows, and each data point is a $d$ dimensional row vector ($d \in \mathbb{N}$), augmented by unity, having a total length of $d + 1$.
    \item $Y$: Regression labels ($Y \in \mathbb{R}^N$), i.e. the dependant variable in linear regression.
    \item $w$: Regression weights to be learned, $w \in \mathbb{R}^{d + 1}$.
\end{itemize}

In Figure \ref{fig:regression-explained}, the red dots represent the regression training data and the blue line represents the best fit curve for the given training data. 
With reference to Figure \ref{fig:regression-explained}, the regression problem can be stated as follows:
\begin{align}
    \min_{w \in \mathbb{R}^{d+1}} \ E(w) &= || Xw - Y ||^2 \label{eq:regression}
\end{align}

where, $E(w)$ is the Euclidean error function.
The regression problem is one of the few machine learning problems which has an analytical solution, given by:
\begin{align}
    w = (X^T X)^{-1} X^T Y
\end{align}

If the inverse of $X^T X$ does not exist, the pseudo inverse is computed.
The time complexity of linear regression is known to be $\mathcal{O}(N d^2)$.

\section{Formulation for Adiabatic Quantum Computers}
\label{sec:formulation}

Adiabatic quantum computers are adept at approximately solving QUBO problems, which are NP-hard, and defined as:
\begin{align}
    \min_{z \in \mathbb{B}^M} z^T A z + z^T b \label{eq:qubo}
\end{align}

where,
$z \in \mathbb{B}^M$ is the binary decision vector ($M \in \mathbb{N}$);
$A \in \mathbb{R}^{M \times M}$ is the QUBO matrix; and,
$b \in \mathbb{R}^M$ is the QUBO vector.
In order to solve on adiabatic quantum computers, the regression problem needs to be converted into a QUBO problem.
We start by rewriting Problem \ref{eq:regression} as follows:
\begin{align}
    \min_{w \in \mathbb{R}^{d+1}} E(w) 
    &= w^T X^T X w - 2 w^T X^T Y + Y^T Y \label{eq:regression-dotproduct}
\end{align}


Next, we introduce a precision vector $P = [p_1, p_2, \ldots, p_K]^T$, $K \in \mathbb{N}$, which is a constant.
Each entry in $P$ can be an integral power of $2$, and can be both positive or negative.
The precision vector must be sorted.
For example, a precision vector could be: $P = \left[ -2, -1, -\frac{1}{2}, \frac{1}{2}, 1, 2, \right]^T$.
Next, we introduce $K$ binary variables $\hat{w}_{ik}$ for each of the $d+1$ regression weights $w_i$ so that:
\begin{align}
    w_i = \sum_{k=1}^K p_k \hat{w}_{ik} \qquad \forall i = 1, 2, \ldots, d + 1 \label{eq:binarized-regression-weights}
\end{align}

where, $p_k$ denotes the $k^{th}$ entry in the precision vector $P$.
$\hat{w}_{ik}$ can be thought of as a binary decision variable that selects or ignores entries in $P$ depending on whether its value is $1$ or $0$ respectively.
With this formulation, we can have up to $2^K$ unique values for each $w_i$ when $P$ contains only positive values for instance.
However, if $P$ contains negative values as well, then the number of unique attainable values for each $w_{i}$ might be less than $2^K$.
For example, if $P = [-1, -\frac{1}{2}, \frac{1}{2}, 1]$, then only the following seven distinct values can be attained: $\{-\frac{3}{2}, -1, -\frac{1}{2}, 0, \frac{1}{2}, 1, \frac{3}{2}\}$.
Next, we rewrite Equation \ref{eq:binarized-regression-weights} in a matrix form as follows:
\begin{align}
    w &= \mathcal{P} \hat{w} \label{eq:binarized-weights-matrix-form}
\end{align}

where,
$\mathcal{P} = I_{d+1} \otimes P^T$ is the $(d+1) \times K(d+1)$ precision matrix obtained by taking the Kronecker product of identity matrix ($I_{d+1}$) with transpose of precision vector ($P$); and,
$\hat{w} = [\hat{w}_{11}, \ldots, \hat{w}_{1K}, \hat{w}_{21}, \ldots, \hat{w}_{2K}, \ldots, \hat{w}_{(d+1)1}, \ldots, \hat{w}_{(d+1)K}]^T$ is the vector containing all $(d+1) K$ binary variables introduced in Equation \ref{eq:binarized-regression-weights}.
These steps are taken for mathematical convenience.
Now that we have expressed $w$ in terms of binary variables $\hat{w}$ and precision matrix $\mathcal{P}$, we can substitute the value of $w$ from Equation \ref{eq:binarized-weights-matrix-form} into Equation \ref{eq:regression-dotproduct}, and convert the regression problem into a QUBO problem as follows:
\begin{align}
    \min_{\hat{w} \in \mathbb{B}^{(d+1)K}} E(\hat{w}) &= \hat{w}^T \mathcal{P}^T X^T X \mathcal{P} \hat{w} - 2 \hat{w}^T \mathcal{P}^T X^T Y \label{eq:regression-to-qubo}
\end{align}


Note that we left out the last term ($Y^T Y$) from Equation \ref{eq:regression-dotproduct} because it is a constant scalar and does not affect the optimal solution of the unconstrained optimization problem.
Also, note that Equation \ref{eq:regression-to-qubo} is identical to Equation \ref{eq:qubo}, with $M = (d+1) K$, $z = \hat{w}$, $A = \mathcal{P}^T X^T X \mathcal{P}$ and $b = - 2 \mathcal{P}^T X^T Y$.
Thus, Equation \ref{eq:regression-to-qubo} is a QUBO problem and can be solved on adiabatic quantum computers.

\section{Analysis}
\label{sec:analysis}

\subsection{Theoretical Analysis}
\label{sub:theoretical}

The regression problem (Problem \ref{eq:regression}) has $\mathcal{O}(N d)$ data ($X$ and $Y$) and $\mathcal{O}(d)$ weights ($w$), which is the same for Problem \ref{eq:regression-to-qubo}.
While converting Problem \ref{eq:regression} to Problem \ref{eq:regression-to-qubo}, we introduced $K$ binary variables for each of the $d+1$ weights.
So, we have $\mathcal{O}(d K)$ variables in Equation \ref{eq:regression-to-qubo}, which translates to quadratic qubit footprint ($\mathcal{O}(K^2 d^2)$) using an efficient embedding algorithm like \cite{date2019efficiently}. 
Embedding is the process of mapping logical QUBO variables to qubits on the hardware, and is challenging because inter-qubit connectivity on the hardware is extremely limited.
As mentioned in Section \ref{sec:background}, solving the regression problem (Equation \ref{eq:regression}) takes $\mathcal{O}(N d^2)$ time.
From Equation \ref{eq:regression-to-qubo}, we can infer that the QUBO formulation takes $\mathcal{O}(N d^2 K^2)$ time.
Obtaining the solution on adiabatic quantum computers depends on the annealing time, which is not $\mathcal{O}(1)$ in general, but can be treated as $\mathcal{O}(1)$ for all practical purposes.
So, the total time to convert and solve a linear regression problem on adiabatic quantum computer would be $\mathcal{O}(N d^2 K^2)$.

It is clear that this running time is worse than its classical counterpart ($\mathcal{O}(N d^2)$).
But, the above analysis assumes that $K$, which is the length of the precision vector, is a variable.
On classical computers, the precision is fixed, for example, $32$-bit or $64$-bit precision.
We can analogously fix the precision for quantum computers, and treat $K$ as a constant.
The resulting qubit footprint would be $\mathcal{O}(d^2)$, and the time complexity would be $\mathcal{O}(N d^2)$, which is equivalent to the classical algorithm.

\subsection{Empirical Analysis}
\label{sub:empirical}

\subsubsection{Methodology and Performance Metrics}
We test our quantum approach for regression using the D-Wave 2000Q adiabatic quantum computer and compare it to a classical approach using the Scikit-learn library in Python.
The Scikit-learn library is widely used for machine learning tasks like linear regression, support vector machines, K-nearest neighbors, K-means clustering etc.
We use two performance metrics for this comparison: 
(i) Regerssion error (Equation \ref{eq:regression}); and, 
(ii) Total computation time.
For D-Wave, the total computation time is comprised of the preprocessing time and the annealing time.
The preprocessing time refers to converting the regression problem into QUBO problem and embedding it for the D-Wave hardware using our embedding algorithm from \cite{date2019efficiently}.
It must be noted that while working with D-Wave, there is a significant amount of time spent on sending a problem to the D-Wave servers, and receiving the solution back, which we refer to as network overheads.
Although we report network overheads in Tables \ref{tab:regression-scaling-N} and \ref{tab:regression-scaling-d} for information purposes, we do not plot them in Figures \ref{fig:regression-scaling-N} and \ref{fig:regression-scaling-d} and exclude them from our algorithm's run time.
This is because the network overheads are determined by factors like physical proximity of a user to D-Wave servers, network connectivity etc., which are neither in our control nor exclusive to our algorithm.
In this paper, each quantum annealing operation is performed $1,000$ times and only the ground state solution is used.
The value of $1,000$ was seen to yield most reliable results.

\subsubsection{Data Generation}
All data in this study, including the ground truth weights were synthetically generated, uniformly at random to curb any biases.
We also injected noise into the data in order to compare robustness of both approaches and to emulate noisy nature of real world data.
The precision vector $P$ is constant across all our experiments, and the ground truth weights can be attained using the entries of $P$.
We tried using benchmark datasets for regression like body fat, housing and pyrim \cite{gao2015sample}, but couldn't generate any meaningful results because of the limitations imposed by the hardware architecture of the D-Wave 2000Q.
These benchmark datasets require at least 16-bit precision and have several features.
The D-Wave machine was too small to accommodate the QUBO problems that stem from these datasets.

\subsubsection{Hardware Configuration}
Preprocessing for our quantum approach and entire classical approach were run on a machine with 3.6 GHz 8-core Intel i9 processor and 64 GB 2,666 MHz DDR4 memory.
The quantum approach also used the D-Wave 2000Q quantum computer, which had $2,048$ qubits and about $5,600$ inter-qubit connections.

\subsubsection{Comparing Regression Error}
\begin{table}[t!]
    \centering
    \caption{Comparing Regression Error}
    \begin{tabular}{m{0.25\textwidth} m{0.08\textwidth} m{0.07\textwidth}}
        \noalign{\smallskip} \hline \noalign{\smallskip}
                               Experimental Runs Where & Scikit-learn Error    & D-Wave Error \\
        \noalign{\smallskip} \hline \noalign{\smallskip}
        D-Wave fit the data ($80\%$ runs)          & 5.0261                & 5.0362  \\
        D-Wave did not fit the data ($20\%$ runs)        & 4.8188 & 15.0657 \\
        Overall & 4.9846                & 7.0421  \\
        \noalign{\smallskip} \hline \noalign{\smallskip}
    \end{tabular}
    \label{tab:regression-error}
\end{table}
\begin{figure}[t!]
    \centering
    \includegraphics[scale=0.5]{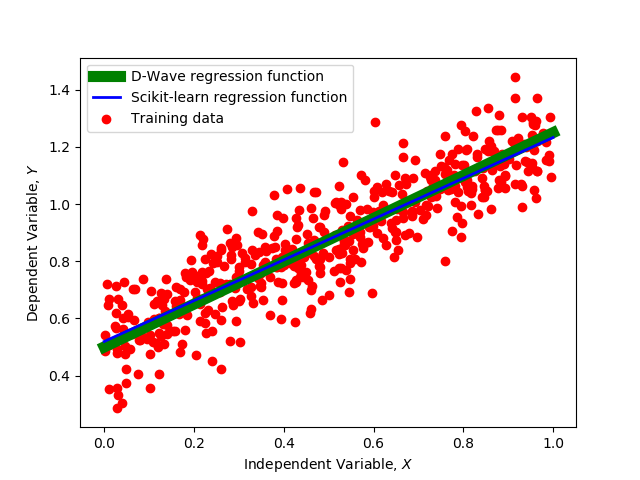}
    \caption{Comparison of regression curves fit by Scikit-learn (blue) and D-Wave (green) on synthetic data (red circles). X-axis shows the independent variable and Y-axis shows the dependent variable. Both curves closely resemble each other.}
    \label{fig:regression-comparison}
\end{figure}

We compute regression error (Equation \ref{eq:regression}) for our quantum approach using D-Wave 2000Q and compare it to the classical approach using Scikit-learn in Table \ref{tab:regression-error}.
We report mean errors over $100$ identical experimental runs to assess recovery rate of the D-Wave machine.
The ground truth regression weights for all the experimental runs were $[0.5, 0.75]$ and the precision vector used was $P = [0.25, 0.5]$.
In each experimental run, we generate regression training data synthetically based on the ground truth regression weights, feed this data to both classical and quantum approaches, and compute the regression error.
We observe that the D-Wave approach fit the regression training data about $80\%$ of the time with a mean error of $5.0362$. 
The mean Scikit-learn error for these runs was $5.0261$.
While both errors are in the same ballpark, the Scikit-learn error is slightly lower than D-Wave because of higher precision on 64-bit machine.
Within the 2-bit precision allowed by the precision vector $P$, D-Wave was seen to find the best possible solution.
An illustration of this is shown in Figure \ref{fig:regression-comparison}, where regression data is shown by red dots, Scikit-learn function is shown by blue line and D-Wave function is shown by green line.
Both Scikit-learn and D-Wave functions closely resemble each other, and are able to fit the data.
In the case where D-Wave did not fit the regression data ($20\%$ of the time), mean D-Wave error was $15.0657$.
Mean Scikit-learn error for these runs was $4.8188$.
On an average, the Hamming distance (number of bit-flips) between D-Wave solutions and the ground truth solutions was two across the four binary variables in this problem.
The reason for this discrepancy is ingrained in the hardware of the D-Wave machine, which is known to produce faulty results when inter-qubit connections break during quantum annealing \cite{king2014algorithm}.
Overall, mean errors for Scikit-learn and D-Wave were $4.9846$ and $7.0421$ respectively.

\subsubsection{Scalability with Number of Datapoints ($N$)}
\begin{table*}[h!]
    \centering
    \caption{Scalability with Number of Datapoints ($N$)}
    \begin{tabular}{m{0.07\textwidth} m{0.15\textwidth} m{0.125\textwidth} m{0.12\textwidth} m{0.125\textwidth} m{0.13\textwidth}}
        \noalign{\smallskip} \hline \noalign{\smallskip}
        Number of Datapoints ($N$) & \textbf{Scikit-learn Time (ms)} & D-Wave Preprocessing Time (ms) & D-Wave Annealing Time (ms) & \textbf{D-Wave Compute Time (Preprocess + Anneal) (ms)} & D-Wave Network Overheads (ms) \\
        \noalign{\smallskip} \hline \noalign{\smallskip}
        512         & \textbf{0.7976 $\pm$ 0.0780}       & 0.2594 $\pm$ 0.0437   & 12.5151 $\pm$ 0.0186   & \textbf{12.7744 $\pm$ 0.0461}     & 703.5815 $\pm$ 54.9066    \\
        1,024       & \textbf{0.8274 $\pm$ 0.0957}       & 0.2543 $\pm$ 0.0261   & 12.5143 $\pm$ 0.0146   & \textbf{12.7686 $\pm$ 0.0309}     & 703.0153 $\pm$ 34.5316    \\
        2,048       & \textbf{0.8677 $\pm$ 0.0801}       & 0.2997 $\pm$ 0.0470   & 12.5152 $\pm$ 0.0105   & \textbf{12.8149 $\pm$ 0.0481}     & 703.9943 $\pm$ 33.6994    \\
        4,096       & \textbf{0.9259 $\pm$ 0.0890}       & 0.3284 $\pm$ 0.0337   & 12.5192 $\pm$ 0.0063   & \textbf{12.8475 $\pm$ 0.0336}     & 689.5000 $\pm$ 35.5189    \\
        \noalign{\smallskip} \hline \noalign{\smallskip}
        8,192       & \textbf{1.0851 $\pm$ 0.0818}       & 0.3635 $\pm$ 0.1089   & 12.5205 $\pm$ 0.0036   & \textbf{12.8840 $\pm$ 0.1088}     & 704.1441 $\pm$ 33.4093    \\ 
        16,384      & \textbf{1.2458 $\pm$ 0.0895}       & 0.3041 $\pm$ 0.1913   & 12.5166 $\pm$ 0.0070   & \textbf{12.8207 $\pm$ 0.1904}     & 716.4246 $\pm$ 45.7286    \\
        32,768      & \textbf{1.6180 $\pm$ 0.0975}       & 0.4304 $\pm$ 0.2380   & 12.5129 $\pm$ 0.0079   & \textbf{12.9433 $\pm$ 0.2368}     & 712.0551 $\pm$ 35.4758    \\
        65,536      & \textbf{2.7692 $\pm$ 0.1485}       & 0.5584 $\pm$ 0.3751   & 12.5186 $\pm$ 0.0080   & \textbf{13.0770 $\pm$ 0.3760}     & 718.3913 $\pm$ 40.5731    \\
        \noalign{\smallskip} \hline \noalign{\smallskip}
        131,072     & \textbf{4.8113 $\pm$ 0.2198}       & 1.1546 $\pm$ 0.6897   & 12.5149 $\pm$ 0.0112   & \textbf{13.6695 $\pm$ 0.6906}     & 702.8292 $\pm$ 38.7911    \\
        262,144     & \textbf{9.9080 $\pm$ 0.6120}       & 2.7862 $\pm$ 1.0094   & 12.5155 $\pm$ 0.0076   & \textbf{15.3017 $\pm$ 1.0088}     & 711.5130 $\pm$ 37.2957    \\
        524,288     & \textbf{19.5373 $\pm$ 1.0212}      & 5.1193 $\pm$ 0.3992   & 12.5166 $\pm$ 0.0030   & \textbf{17.6358 $\pm$ 0.3983}     & 709.6294 $\pm$ 39.2782    \\
        1,048,576   & \textbf{37.3581 $\pm$ 1.8984}      & 10.4900 $\pm$ 0.6307  & 12.5167 $\pm$ 0.0024   & \textbf{23.0067 $\pm$ 0.6307}     & 707.4266 $\pm$ 38.3336    \\
        \noalign{\smallskip} \hline \noalign{\smallskip}
        2,097,152   & \textbf{73.6735 $\pm$ 3.4312}      & 27.0889 $\pm$ 1.3411  & 12.5175 $\pm$ 0.0025   & \textbf{39.6064 $\pm$ 1.3413}     & 716.9348 $\pm$ 36.1262    \\
        4,194,304   & \textbf{159.1724 $\pm$ 8.2130}     & 55.3273 $\pm$ 3.1763  & 12.5178 $\pm$ 0.0069   & \textbf{67.8451 $\pm$ 3.1759}     & 713.8490 $\pm$ 52.9194    \\
        8,388,608   & \textbf{328.2112 $\pm$ 13.0534}    & 103.6629 $\pm$ 4.5238 & 12.5170 $\pm$ 0.0036   & \textbf{116.1799 $\pm$ 4.5245}    & 718.6187 $\pm$ 41.1655    \\
        16,777,216  & \textbf{635.9468 $\pm$ 20.6696}    & 214.2371 $\pm$ 8.4610 & 12.5202 $\pm$ 0.0076   & \textbf{226.7573 $\pm$ 8.4616}    & 710.6270 $\pm$ 32.7847    \\
        \noalign{\smallskip} \hline \noalign{\smallskip}
    \end{tabular}
    \label{tab:regression-scaling-N}
\end{table*}

\begin{figure*}[h!]
    \centering
    \begin{subfigure}{0.45\textwidth}
        \centering
        \includegraphics[scale=0.5]{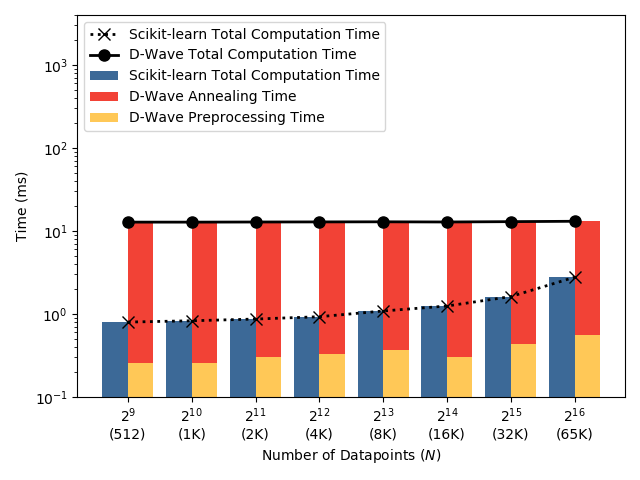}
        \caption{Small number of datapoints ($N$)}
        \label{fig:regression-scaling-N-small}
    \end{subfigure}
    \begin{subfigure}{0.45\textwidth}
        \centering
        \includegraphics[scale=0.5]{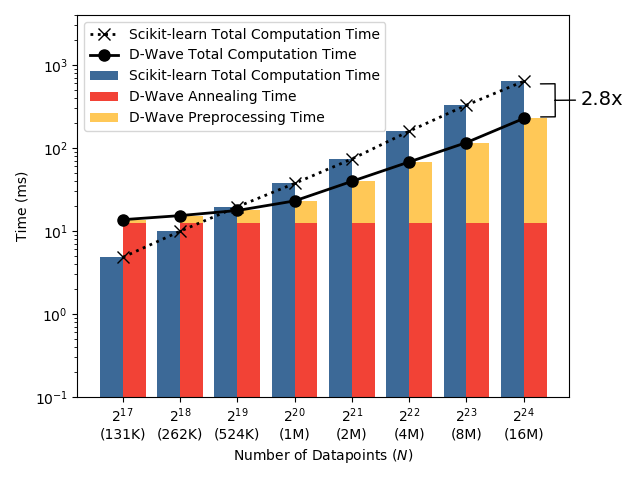}
        \caption{Large number of datapoints ($N$)}
        \label{fig:regression-scaling-N-large}
    \end{subfigure}
    \caption{Scalability comparison of Scikit-learn regression (blue bars and dotted line) and D-Wave regression (yellow and red bars, and bold line). X-axis shows number of datapoints in the training set ($N$), ranging from $2^{9}$ ($512$) to $2^{24}$ ($16$ million) across both figures. Y-axis shows run time milliseconds on a logarithmic scale. In Figure \ref{fig:regression-scaling-N-small}, $N$ varies from $512$ to $65,536$. In Figure \ref{fig:regression-scaling-N-large}, $N$ varies from $131,072$ to $16,777,216$. We observe a $2.8 \times$ speedup using D-Wave on the 16 million case in Figure \ref{fig:regression-scaling-N-large}.}
    \label{fig:regression-scaling-N}
\end{figure*}

We perform a scalability study to determine how the run time of our quantum approach as well as the classical approach changes as the size of regression dataset increases from $512$ datapoints to over $16$ million datapoints.
We report the mean and standard deviation across 60 runs in Table \ref{tab:regression-scaling-N} and fix the number of features ($d+1$) at $2$.
The scalability results are presented in Figure \ref{fig:regression-scaling-N} where the logarithmic X-axis denotes number of datapoints ($N$), the logarithmic Y-axis denotes the time in milliseconds, the blue bars denote total Scikit-learn time, the yellow bars denote D-Wave preprocessing time, and the red bars denote D-Wave annealing time.
We observe that when number of datapoints is small ($N \le 262,144$), Scikit-learn performs faster than D-Wave.
In this case, D-Wave compute time is dominated by annealing time and the preprocessing time is minimal.
When the number of datapoints is large ($N \ge 524,288$), D-Wave performs faster than Scikit-learn.
In this case, D-Wave compute time is dominated by the preprocessing time and the annealing time is minimal.
The run times for the two approaches are comparable when $N$ equals $524,288$ datapoints.
When $N$ equals $16,777,216$, we observe that the quantum approach is $2.8\times$ faster than the classical approach.
Furthermore, we also notice that D-Wave annealing time is essentially constant, and preprocessing time is always less than Scikit-learn time.
This is attributed to efficiently converting regression problem into QUBO problem as described in this paper, and efficiently generating an embedding using our embedding algorithm \cite{date2019efficiently}.
The quantum approach seems to outperform the classical approach on larger datasets.

\subsubsection{Scalability with Number of Features ($d+1$)}
\begin{table*}[h!]
    \centering
    \caption{Scalability with Number of Features ($d+1$)}
    \begin{tabular}{m{0.07\textwidth} m{0.15\textwidth} m{0.125\textwidth} m{0.12\textwidth} m{0.125\textwidth} m{0.16\textwidth}}
        \noalign{\smallskip} \hline \noalign{\smallskip}
        Number of Features ($d+1$) & \textbf{Scikit-learn Time (ms)} & D-Wave Preprocessing Time (ms) & D-Wave Annealing Time (ms) & \textbf{D-Wave Compute Time (Preprocess + Anneal) (ms)} & D-Wave Network Overheads (ms) \\
        \noalign{\smallskip} \hline \noalign{\smallskip}
        2   & \textbf{20.6123 $\pm$ 1.2042}   & 5.1378 $\pm$ 0.3802      & 12.5076 $\pm$ 0.0007  & \textbf{17.6454 $\pm$ 0.3802}  & 706.1933 $\pm$ 84.0029    \\
        4   & \textbf{30.6010 $\pm$ 1.2382}   & 13.7718 $\pm$ 0.9632     & 12.5247 $\pm$ 0.0009  & \textbf{26.2965 $\pm$ 0.9632}  & 754.6531 $\pm$ 67.8583    \\ 
        6   & \textbf{46.8912 $\pm$ 1.6430}   & 21.4310 $\pm$ 1.7784     & 12.5450 $\pm$ 0.0008  & \textbf{33.9759 $\pm$ 1.7783}  & 756.5598 $\pm$ 66.0033    \\
        8   & \textbf{68.4019 $\pm$ 3.9914}   & 28.9740 $\pm$ 2.0591     & 12.5659 $\pm$ 0.0006  & \textbf{41.5398 $\pm$ 2.0590}  & 715.7661 $\pm$ 61.4712    \\
        \noalign{\smallskip} \hline \noalign{\smallskip}
        10  & \textbf{93.9764 $\pm$ 1.7518}   & 35.6321 $\pm$ 2.2202     & 12.5935 $\pm$ 0.0010  & \textbf{48.2257 $\pm$ 2.2203}  & 761.3451 $\pm$ 60.9483    \\
        12  & \textbf{118.1701 $\pm$ 2.0026}  & 42.5206 $\pm$ 2.6595     & 12.6092 $\pm$ 0.0012  & \textbf{55.1297 $\pm$ 2.6596}  & 781.4883 $\pm$ 91.1495    \\
        14  & \textbf{145.6177 $\pm$ 1.8870}  & 52.2676 $\pm$ 3.2121     & 12.6140 $\pm$ 0.0008  & \textbf{64.8816 $\pm$ 3.2120}  & 844.3496 $\pm$ 107.1684   \\
        16  & \textbf{175.5792 $\pm$ 2.4876}  & 60.6022 $\pm$ 3.7414     & 12.6195 $\pm$ 0.0010  & \textbf{73.2217 $\pm$ 3.7415}  & 877.4846 $\pm$ 103.0307   \\
        \noalign{\smallskip} \hline \noalign{\smallskip}
        18  & \textbf{213.1724 $\pm$ 2.7332}  & 65.1949 $\pm$ 3.6451     & 12.6222 $\pm$ 0.0004  & \textbf{77.8170 $\pm$ 3.6451}  & 791.8038 $\pm$ 105.9935   \\
        20  & \textbf{236.3750 $\pm$ 4.6308}  & 74.9983 $\pm$ 3.9706     & 12.6212 $\pm$ 0.0004  & \textbf{87.6194 $\pm$ 3.9706}  & 920.5470 $\pm$ 55.5933    \\
        22  & \textbf{257.6503 $\pm$ 5.2920}  & 80.3314 $\pm$ 4.9133     & 12.6215 $\pm$ 0.0005  & \textbf{92.9529 $\pm$ 4.9133}  & 779.6944 $\pm$ 90.0467    \\
        24  & \textbf{281.3093 $\pm$ 4.0617}  & 83.6467 $\pm$ 3.6455     & 12.6211 $\pm$ 0.0007  & \textbf{96.2678 $\pm$ 3.6457}  & 847.5243 $\pm$ 113.3988   \\
        \noalign{\smallskip} \hline \noalign{\smallskip}
        26  & \textbf{313.2982 $\pm$ 3.7929}  & 89.3653 $\pm$ 2.9930     & 12.6197 $\pm$ 0.0008  & \textbf{101.9850 $\pm$ 2.9931} & 804.1284 $\pm$ 91.5301    \\
        28  & \textbf{343.8831 $\pm$ 4.3417}  & 98.9982 $\pm$ 4.5461     & 12.6174 $\pm$ 0.0010  & \textbf{111.6156 $\pm$ 4.5462} & 900.9338 $\pm$ 50.4548    \\
        30  & \textbf{379.3123 $\pm$ 4.7932}  & 108.6154 $\pm$ 5.0667    & 12.6093 $\pm$ 0.0007  & \textbf{121.2247 $\pm$ 5.0667} & 1709.0392 $\pm$ 7766.5828 \\
        32  & \textbf{360.5327 $\pm$ 9.4234}  & 116.3282 $\pm$ 5.2455    & 12.5901 $\pm$ 0.0004  & \textbf{128.9182 $\pm$ 5.2455} & 706.8372 $\pm$ 73.4864    \\
        \noalign{\smallskip} \hline \noalign{\smallskip}
    \end{tabular}
    \label{tab:regression-scaling-d}
\end{table*}

\begin{figure}[h!]
    \centering
    \includegraphics[scale=0.5]{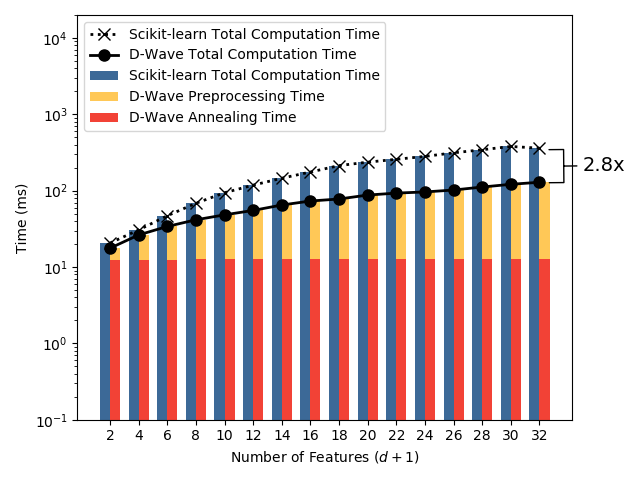}
    \caption{Scalability of Scikit-learn regression (blue bars and dotted line) and D-Wave regression (yellow and red bars, and bold line). X-axis shows number of features in the training set ($d+1$), ranging from $2$ to $32$. The Y-axis shows run time in milliseconds on a logarithmic scale. We observe a $2.8 \times$ speedup using D-Wave when $(d+1)$ equals $32$.}
    \label{fig:regression-scaling-d}
\end{figure}

We assess the scalability with respect to the number of features ($d+1$) as well.
To eliminate the effect of number of datapoints, we fix $N$ at $524,288$ datapoints because from Table \ref{tab:regression-scaling-N} and Figure \ref{fig:regression-scaling-N}, the run times of both quantum and classical approaches are comparable at this value.
The results are presented in Table \ref{tab:regression-scaling-d} and Figure \ref{fig:regression-scaling-d}, where we vary the number of features ($d+1$) from $2$ to $32$.
In Table \ref{tab:regression-scaling-d}, we report the mean and standard deviation over $60$ runs for each experimental configuration.
In Figure \ref{fig:regression-scaling-d}, the X-axis shows number of features ($d+1)$, the logarithmic Y-axis shows run time in milliseconds, the blue bars denote total Scikit-learn times, the yellow bars denote D-Wave preprocessing times and the red bars denote D-Wave annealing times.
We observe that D-Wave performs faster than Scikit-learn for all values of $d+1$, and attains $2.8\times$ speedup when $d+1$ equals 32.
We also observe that D-Wave run time is dominated by preprocessing time for almost all values of $d+1$, but is always less than Scikit-learn.
This is attributed to efficient conversion of regression into QUBO as outlined in this paper, and use of our efficient embedding algorithm \cite{date2019efficiently}.
Lastly, we notice that the D-Wave annealing time is essentially constant across all values of $d+1$. 
As the number of features ($d+1$) increase, the quantum approach is seen to perform faster than the classical approach.

\subsection{Discussion}

We first address why it is possible to scale $N$ to over $16$ million datapoints, but not possible to scale $d+1$ over $32$ features.
In Section \ref{sec:formulation}, we show that the the qubit footprint (number of qubits used) of our formulation is $\mathcal{O}(d^2)$, and is independent of $N$, allowing us to scale $N$ to over $16$ million.
We refrained from scaling $N$ to larger values because we believe $16$ million is a large enough value to convey the crux of this work---quantum computers can be valuable for solving machine learning problems like linear regression, especially on larger sized problems.
We are limited to values of $d+1$ that are smaller than $32$ because the qubit footprint depends on $d$.
The size of the largest problem with all-to-all connectivity that can be accommodated on D-Wave 2000Q is $64$, i.e. a QUBO problem having $64$ variables.
This is determined by the hardware architecture.
Based on our formulation, the size of the regression QUBO problem is $(d+1) K$. 
So, problems for which $(d+1) K \le 64$ can be accommodated on the D-Wave machine.
In our experimental runs, we fixed $K$ as $2$, and therefore, must have $(d+1) \le 32$.
This limitation stems from the number of qubits and inter-qubit connectivity available on today's quantum computers, and will be improved in future quantum computers, which are sought to be bigger and more reliable than the current machines.
For instance, the next generation D-Wave machines would have 5,000 qubits and would support more inter-qubit connections \cite{boothby2020next,dattani2019pegasus}.

Secondly, we would like to reiterate that D-Wave was seen to produce accurate results about $80\%$ of the time during our empirical analysis, which is better than $50\%$ recovery rate previously observed by Chang et al. \cite{chang2019quantum}. 
This could be attributed to hardware and software improvements made by D-Wave to their systems.
During the remaining $20\%$ of the time, the inter-qubit connections on the hardware had a tendency to break, resulting in inferior solutions.
This became increasingly prevalent on larger problems, which use large number of qubits.
This hardware issue is ubiquitous across noisy intermediate-scale quantum (NISQ) computers and is expected to get better in the future as improved engineering solutions are deployed for building these machines.

Lastly, we would like to emphasize the algorithmic gains that could be realized by using our quantum approach for linear regression.
In our empirical analysis, we observed that the quantum approach essentially had constant annealing time and that the preprocessing time was always less than the run time of the classical approach.
For embedding QUBO problems onto the D-Wave hardware, we tried using D-Wave's embedding algorithm, but got significantly inferior results. 
All results in this paper use our embedding algorithm, which is described in \cite{date2019efficiently}.
Our quantum approach performed faster than the classical approach on increasingly large values of number of datapoints ($N$) as well as number of features ($d$).
With quantum computers becoming less prone to error-correction in the future, it might be beneficial to use a quantum approach for linear regression, especially on larger problems.

\section{Data Availability}
The data that support the findings of this study are available from the corresponding author upon reasonable request.

\section{Conclusion}
\label{sec:conclusion}

Training machine learning models for real world applications is time intensive and can even take a few months in some cases.
Generally, training a machine learning model is equivalent to solving an optimization problem over a well defined error function.
Quantum computers are known to be good at (approximately) solving hard optimization problems and offer a compelling alternative for training machine learning models.
In this paper, we propose an adiabatic quantum computing approach for training linear regression models, which is a statistical machine learning technique.
We analyze our quantum approach theoretically, compare it to current classical approaches, and show that the time complexity for both these approaches is equivalent.
Next, we test our quantum approach using the D-Wave 2000Q adiabatic quantum computer and compare it to a classical approach using the Scikit-learn library in Python.
We demonstrate that the quantum approach performs at par with the classical approach on the regression error metric, and attains $2.8 \times$ speedup over the classical approach on larger (synthetically generated) datasets.

Continuing along this line of research, we would like to test our approach on real world datasets that can be accommodated on today's quantum computers.
We would also like to extend our quantum approach to variants of linear regression that use kernel methods.
Finally, we would like to explore the use of quantum computers for training other machine learning models like Support Vector Machines (SVM), Deep Neural Networks (DNN), Generative Adversarial Networks (GAN) etc.

\section{Competing Interests}
The authors declare that there are no competing interests.

\section{Author Contribution}
Prasanna Date ideated the work, formulated the regression problem as a quadratic unconstrained binary optimization problem, performed theoretical and empirical analysis and wrote the paper.
Thomas Potok mentored Prasanna during the course of this work and helped write the paper.

\bibliographystyle{IEEEtran}
\bibliography{reference.bib}

\end{document}